\documentclass[letterpaper]{article}
\usepackage{spconf,amsmath,graphicx}
\usepackage[T1]{fontenc}
\usepackage{booktabs}
\usepackage[table]{xcolor}
\usepackage{placeins}
\usepackage{needspace}

\definecolor{TblHead}{RGB}{239,242,245}
\definecolor{PanelA}{RGB}{220,233,244}
\definecolor{PanelARow}{RGB}{247,250,252}
\definecolor{PanelB}{RGB}{248,232,207}
\definecolor{PanelBRow}{RGB}{253,248,240}
\definecolor{KeyRow}{RGB}{241,245,248}

\title{\vspace{4pt}\fontsize{14}{16}\selectfont RETHINKING LENGTH-BASED TRAINING: BATCH COMPOSITION\\ AND LOSS NORMALIZATION IN SPEECH TOKEN LANGUAGE MODELS}
\name{Hongjin Song$^{1}$, Runwu Shi$^{2}$, Weiqiao Shan$^{3}$, Jiale Luo$^{4}$, Yujin Wang$^{5}$, Yifei Wu$^{6}$, Chunxiang Jin$^{6,*}$}
\address{$^{1}$Beijing Institute of Technology, Zhuhai; $^{2}$Institute of Science Tokyo; $^{3}$Northeastern University\\
$^{4}$Sichuan University; $^{5}$Wuhan University; $^{6}$Ant Group}

\begin{document}
\ninept
\maketitle
\begin{abstract}
Short-to-long training is a simple curriculum for speech models, but its gains
can be difficult to interpret. In speech token language models, length-based
training can change the shuffle policy, batch composition, token retention,
and token weights under batch-mean loss. We disentangle these factors through
matched comparisons. In the tested settings, short-to-long ordering shows no
independent benefit when batch composition and token exposure are fixed.
First-epoch grouping lowers perplexity for Mimi under batch-mean loss, but
this gain is not observed under token-balanced loss. The cross-tokenizer results are
consistent with a link between chunk-length variation and token weighting.
This work provides a systematic analysis protocol for studying length-based
training in variable-length speech models.

\end{abstract}
\section{Introduction}
\label{sec:intro}
Autoregressive modeling of discrete speech tokens has received increasing
attention in speech generation and speech language modeling
\cite{ji2025wavtokenizer,ji2025languagecodec,dellalibera2025focalcodec}.
The length of a speech-token sequence depends on both utterance duration and
tokenizer rate. This affects batch construction and the number of valid tokens
in each update. SortaGrad presents shorter utterances first in the first epoch
and then returns to random minibatch order~\cite{amodei2016deepspeech2}.
Recent work studies curriculum schedules with token budgets~\cite{zhang2026beyond},
difficulty-based token-loss weighting~\cite{jia2026curriculum}, and within-batch
diversity~\cite{dai2026organization}. Sequence length also matters in speech
data selection~\cite{whetten2026selection}.

Document packing also changes training: Best-fit Packing reduces document
fragmentation while preserving training efficiency~\cite{ding2024truncation}.
Here, length sorting changes batch membership and, under min-length
truncation, which targets are retained. Per-batch mean loss also assigns
larger coefficients to token losses in batches with fewer valid targets.
A sorted-versus-random comparison can therefore mix presentation order with
changes in target exposure and loss normalization.

\begin{figure*}[t]
\centering
\includegraphics[width=\textwidth]{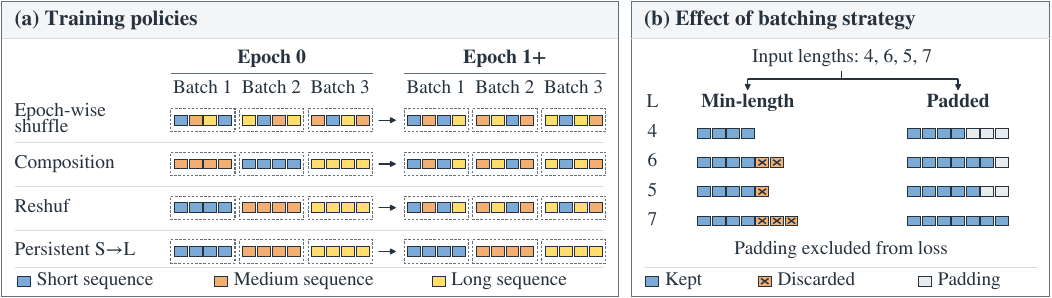}
\caption{Overview of the factors changed by length-based training. (a) Training
policies across epochs. Composition and Reshuf use the same first-epoch length
grouping but differ in batch order; both return to epoch-wise shuffling
afterward, while Persistent S$\rightarrow$L repeats grouping and ascending
order. (b) For an example variable-length batch, min-length batching discards
valid tokens beyond the shortest sequence, whereas padded batching keeps all
valid tokens and masks padding in the loss.}
\label{fig:overview}
\end{figure*}

We ask whether the gains from short-to-long training come from presentation
order or from the changes in batching that accompany length sorting. Our
fixed-batch comparisons isolate presentation order, and a normalization
intervention tests the remaining first-epoch grouping effect. The study uses an 87M-parameter
Transformer on LibriSpeech train-clean-100, with Mimi for the main analysis
and EnCodec and SpeechTokenizer for cross-tokenizer comparisons.

With padded evaluation and fixed batch membership, sorted orders do not
improve perplexity in the tested settings. First-epoch grouping lowers PPL
for Mimi under batch-mean loss, but this gain is not observed under
token-balanced loss. The normalization change has little effect on Shuffle
but raises PPL for Composition. The controlled comparisons separate
presentation order from batch construction and loss normalization.

\section{Method}
\label{sec:method}

\subsection{Training factors}
\label{sec:factors}

Let $\mathbf{x}_i=(x_{i,1},\ldots,x_{i,c_i})$ be a stored token chunk.
Its next-token target length is $\ell_i=c_i-1\leq M$, with $M=256$.
Length-based training changes the factors below.

\textbf{Shuffle policy.}
A random permutation can be sampled once and reused, or sampled again at every
epoch. We call these settings \emph{fixed random order} and
\emph{epoch-wise shuffle}. They match the SingleShuffle/RandomShuffle
distinction in optimization work~\cite{ahn2020shuffling}. Epoch-wise shuffling
also changes which sequences share a batch across epochs.

\textbf{Batch composition and order.}
Length sorting affects both which sequences appear in the same batch and the
order in which batches are processed. We distinguish these two effects.
\emph{Length grouping} forms batches from sequences with similar lengths,
while \emph{short-to-long order} presents the resulting batches in ascending
length order. The batch order can instead be randomized while preserving the
same length-grouped composition.

\textbf{Token retention and capacity use.}
For a processed batch $\mathcal B$, let $T_{\mathcal B}=\sum_i\ell_i$ before truncation.
Min-length batching retains $K_{\mathcal B}=|\mathcal B|\min_i\ell_i$
targets; padded batching retains all $T_{\mathcal B}$ targets. Across batches,
\begin{equation}
R_{\rm ret}=\frac{\sum_{\mathcal B}K_{\mathcal B}}{\sum_{\mathcal B}T_{\mathcal B}},
\label{eq:retention}
\end{equation}
\begin{equation}
U_{\rm cap}=\frac{\sum_{\mathcal B}K_{\mathcal B}}{M\sum_{\mathcal B}|\mathcal B|}.
\label{eq:capacity}
\end{equation}
$R_{\rm ret}$ is true target retention; $U_{\rm cap}$ measures use of fixed
maximum capacity. Near-equal short chunks can have high retention but low
capacity use. Table~\ref{tab:batchstats} reports $U_{\rm cap}$.

\subsection{Training configurations}
\label{sec:controls}

The factors above are separated through the training configurations in
Table~\ref{tab:settings}. Here, ``grouped'' denotes length-homogeneous
batches.

\emph{Composition} and \emph{Reshuf} use the same length-grouped batches in
the first epoch. Composition presents these batches in random order; Reshuf
presents them from short to long. Both return to epoch-wise shuffling
afterward. Their comparison isolates first-epoch batch order while keeping
batch composition fixed. Persistent short-to-long training repeats length
grouping and ascending order at every epoch.

\begin{table}[t]
\centering
\caption{Training configurations used in the factorization.}
\label{tab:settings}
\small
\renewcommand{\arraystretch}{1.04}
\setlength{\tabcolsep}{3.1pt}
\begin{tabular}{@{}lccc@{}}
\toprule
\rowcolor{TblHead}
Setting & Ep.~0 batch & Ep.~0 order & Ep.~1+ \\
\midrule
Epoch-wise shuffle & Random  & Random & Shuffle \\
Composition        & Grouped & Random & Shuffle \\
Reshuf             & Grouped & S$\rightarrow$L & Shuffle \\
Persistent S$\rightarrow$L
                   & Grouped & S$\rightarrow$L & S$\rightarrow$L \\
\bottomrule
\end{tabular}
\end{table}

The fixed-batch comparison reuses batch membership and padding masks.
\emph{Batch shuffle} randomizes the batch list at every epoch.
\emph{First-epoch S$\rightarrow$L} sorts batches by mean length in epoch 0
and shuffles them afterward. Two persistent conditions repeat ascending or
descending batch order. The targets within each batch are unchanged.

To examine persistent length grouping separately, we also compare two grouped
settings. In the \emph{static-grouped} setting, the same length-grouped
batches are reused across epochs. In the \emph{dynamic-grouped} setting,
length-grouped batches are reconstructed at each epoch while their global
order remains random. Comparing the two tests whether changes in batch
membership, rather than length grouping itself, account for their training
behavior.
\subsection{Loss normalization}
\label{sec:loss}

For a batch $\mathcal{B}$, let $\mathcal{V}_{\mathcal{B}}$ denote its set of
valid target positions and
$N_{\mathcal{B}}=|\mathcal{V}_{\mathcal{B}}|$. The standard training objective
used in our initial experiments averages token-level cross entropy within
each batch:
\begin{equation}
\mathcal{L}_{\mathrm{mean}}(\mathcal{B})
=
-\frac{1}{N_{\mathcal{B}}}
\sum_{(i,t)\in\mathcal{V}_{\mathcal{B}}}
\log p_{\theta}(x_{i,t}\mid x_{i,<t}).
\label{eq:batchmean}
\end{equation}

Each valid token loss has coefficient $1/N_{\mathcal B}$. Batches with fewer
valid targets assign larger coefficients to individual token losses.

We use \emph{token-balanced} loss as a diagnostic intervention,
\begin{equation}
\mathcal L_{\rm bal}(\mathcal B)=-\frac{1}{Z}
\sum_{(i,t)\in\mathcal V_{\mathcal B}}
\log p_\theta(x_{i,t}\mid x_{i,<t}).
\label{eq:tokenbalanced}
\end{equation}
The fixed constant $Z$ is the mean \emph{stored} token count in the reference
length-grouped batches, before next-token shifting. It is shared by the
compared settings. Every valid target loss has coefficient $1/Z$.
For a fixed batch and model state,
$\mathcal L_{\rm bal}=(N_{\mathcal B}/Z)\mathcal L_{\rm mean}$.
If valid-target counts are constant, the two losses differ only by a constant
scale. Loss coefficients are not ratios of AdamW parameter updates, which
also depend on gradient moment estimates.

\begin{figure*}[!t]
\centering
\includegraphics[width=0.98\textwidth]{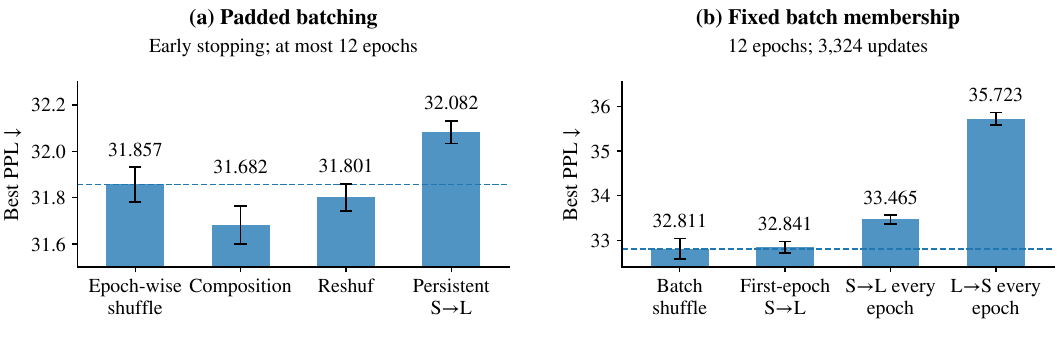}
\caption{Grouping and ordering comparisons on Mimi. (a) Padded policies with
early stopping (Table~\ref{tab:factorization}). (b) Fixed batch membership,
12 epochs, and 3,324 updates (Table~\ref{tab:audits}A). Bars show mean recorded
best PPL over eight seeds; error bars show one standard deviation. Dashed
lines mark each panel's shuffle baseline.}
\label{fig:fair}
\end{figure*}

\section{Experimental Setup}
\label{sec:setup}

\subsection{Data and tokenizers}
\label{sec:data}

The main experiments use LibriSpeech train-clean-100~\cite{panayotov2015librispeech}
with a speaker-disjoint split (split seed 0). Mimi~\cite{defossez2024moshi}
provides the main analysis; EnCodec~\cite{defossez2023encodec} and
SpeechTokenizer~\cite{zhang2024speechtokenizer} provide comparisons.
We use one stream without concatenating codebooks: Mimi uses codebook 0
with one quantizer, EnCodec its first codebook, and SpeechTokenizer its
semantic stream. Token indices are used directly. The vocabulary spans
zero through the largest observed index; padded training adds a PAD symbol.

Chunks contain up to 257 stored tokens, yielding up to 256 next-token targets.
Long utterances use 257-token windows at stride 256 and omit incomplete tails;
short utterances form one chunk. Main runs omit the trailing group of at most
64 chunks; the fixed-batch study keeps every complete training batch.
Padded training uses 257 stored positions. Chunk-length variation is
$\mathrm{CV}(c)=\mathrm{std}(c)/\mathrm{mean}(c)$.

\subsection{Model and training}
\label{sec:training}

Experiments use an 87M-parameter autoregressive Transformer with 12 layers,
12 attention heads, and hidden dimension 768. The batch size is 64.
AdamW uses a peak learning rate of $3\times10^{-4}$, weight decay 0.01,
and a one-cycle schedule with 5\% warmup. Gradients are clipped at norm 1.
Main runs use a maximum of 12 epochs, with early-stopping patience three and
an improvement threshold of 0.02 validation PPL. The fixed-batch study runs
all 12 epochs without early stopping: 3,324 updates in every run.

The main comparisons pair eight initialization seeds and share optimizer and
schedule settings. They match the maximum budget, not the executed updates:
Mimi Shuffle runs take 1,662--1,939 updates, whereas Composition and Reshuf
each take 1,939. All have 277 updates per epoch. The normalization intervention
uses $Z=9217.39$ in both arms, computed before next-token shifting.
For 64-chunk reference batches, the mean target count is $Z-64=9153.39$.

\subsection{Evaluation}
\label{sec:evaluation}

Validation uses consecutive full batches: 32 chunks in the main comparisons
and 64 in the fixed-batch study. Padded evaluation retains their non-padding
targets. Each comparison uses the same validation targets within its study.
Perplexity is computed over the evaluated targets:
\begin{equation}
\mathrm{PPL}
=
\exp\left(
-\frac{1}{N}
\sum_{t=1}^{N}
\log p_{\theta}(x_t \mid x_{<t})
\right).
\end{equation}
We report mean~$\pm$~standard deviation of the recorded best PPL, which is
updated only when PPL improves by more than 0.02. Composition and fixed-batch analyses use two-sided paired
$t$-tests. Table~\ref{tab:factorization} reports raw $p$-values.
Holm correction is applied to the two primary decomposition contrasts and,
separately, the two persistent sorted-order comparisons in Table~\ref{tab:audits}A.
For Table~\ref{tab:losscontrol}, we compute
$I_s=\Delta_{\rm comp,s}^{\rm bal}-\Delta_{\rm comp,s}^{\rm mean}$
within each seed and test its mean against zero using a two-sided one-sample
$t$-test ($n=8$, seven degrees of freedom). Interaction and loss-switch
$p$-values are unadjusted.

\section{Results}
\label{sec:results}

\subsection{Length-based training changes more than order}
\label{sec:baseline}

Figure~\ref{fig:fair} summarizes the padded comparisons on Mimi. Panel (a)
varies grouping and presentation order; panel (b) keeps batch membership
fixed and changes order alone. Composition reaches 31.682 PPL, below 31.857
for epoch-wise shuffle, whereas Persistent S$\rightarrow$L reaches 32.082.
With fixed batches, first-epoch short-to-long ordering changes PPL by only
0.029 ($p=0.81$). The benefit of first-epoch grouping is therefore distinct
from the effect of presenting shorter batches first.

\begin{table}[t]
\centering
\caption{Mimi batch statistics (batch size 64, $M=256$).
Count statistics use stored tokens before next-token shifting.
Batch-max padding is a diagnostic, not the fixed training shape.}
\label{tab:batchstats}
\small
\renewcommand{\arraystretch}{1.04}
\setlength{\tabcolsep}{3.0pt}
\begin{tabular}{@{}lcc@{}}
\toprule
\rowcolor{TblHead}
Metric & Shuffle & Grouped \\
\midrule
Within-batch length CV & 0.3065 & \textbf{0.0025} \\
Within-batch max/min & 5.87 & \textbf{1.007} \\
Batch-max padding & 23.1\% & \textbf{0.4\%} \\
Batch stored-token-count CV & 0.038 & 0.309 \\
Min-length capacity use $U_{\rm cap}$ & 12.6\% & \textbf{55.8\%} \\
\bottomrule
\end{tabular}
\end{table}

Table~\ref{tab:batchstats} shows that grouping makes chunks within a batch
nearly equal in length. Batch-max padding is the mean of
$1-\sum_i c_i/(|\mathcal B|\max_i c_i)$ across batches. These statistics
average nine shuffled batchings (three seeds, three epochs); grouping is
deterministic. $U_{\rm cap}$ uses shuffle seed 1, epoch 1.
The increase from 12.6\% to 55.8\% is greater use of the fixed target capacity,
not a true-retention ratio. Grouping also raises the between-batch stored-token-count
CV from 0.038 to 0.309, linking batch construction to loss normalization.

\subsection{Separating ordering from batch composition}
\label{sec:factorization}

With padded batching, all valid tokens are retained, allowing batch
composition and presentation order to be separated. Table~\ref{tab:factorization}
compares epoch-wise shuffle, Composition, Reshuf, and Persistent
S$\rightarrow$L under this setting.

\begin{table}[t]
\centering
\caption{Mimi factorization ($n=8$). Differences and raw $p$-values are relative to epoch-wise shuffle.}
\label{tab:factorization}
\small
\renewcommand{\arraystretch}{1.04}
\setlength{\tabcolsep}{2.3pt}
\begin{tabular}{@{}lccc@{}}
\toprule
\rowcolor{TblHead}
Setting & Best PPL $\downarrow$ & $\Delta$ PPL & $p$ \\
\midrule
Epoch-wise shuffle & $31.857\pm0.075$ & -- & -- \\
\rowcolor{KeyRow}
Composition & $\mathbf{31.682\pm0.083}$ & $-0.175$ & $3.35{\times}10^{-4}$ \\
Reshuf & $31.801\pm0.058$ & $-0.055$ & $1.08{\times}10^{-2}$ \\
Persistent S$\rightarrow$L & $32.082\pm0.049$ & $+0.225$ & $3.35{\times}10^{-6}$ \\
\bottomrule
\end{tabular}
\end{table}

Let $P_S$, $P_C$, and $P_R$ denote the PPL of Shuffle, Composition, and
Reshuf. The total Reshuf effect can be written as
\begin{equation}
P_R-P_S=\underbrace{(P_C-P_S)}_{\Delta_{\rm comp}}
+\underbrace{(P_R-P_C)}_{\Delta_{\rm order}}.
\label{eq:factorization}
\end{equation}
For Mimi, $\Delta_{\rm comp}=-0.175$, while
$\Delta_{\rm order}=+0.120$. Composition lowers PPL under batch-mean loss.
Adding short-to-long order to the same grouped batches raises PPL by 0.120
($p_{\rm Holm}=6.70\times10^{-4}$), removing part of that gain. Persistent
S$\rightarrow$L is also worse than epoch-wise shuffle. The observed gain is
associated with first-epoch batch composition, not its short-to-long order.

\begin{table}[!t]
\centering
\caption{Mimi order and grouping comparisons ($n=8$).
A: differences from Batch shuffle; Holm correction for the two persistent
sorted orders, raw $p$ for First-epoch S$\rightarrow$L.
B: raw paired tests.}
\label{tab:audits}
\small
\renewcommand{\arraystretch}{1.07}
\setlength{\tabcolsep}{1.55pt}
\begin{tabular}{@{}lccc@{}}
\toprule
\rowcolor{TblHead}
Setting & Best PPL $\downarrow$ & $\Delta$ PPL & $p$ \\
\midrule
\rowcolor{PanelA}
\multicolumn{4}{@{}l}{\textbf{A. Fixed batch membership (order only)}} \\
\rowcolor{PanelARow}
Batch shuffle & $32.811\pm0.230$ & -- & -- \\
\rowcolor{PanelARow}
First-epoch S$\rightarrow$L & $32.841\pm0.130$ & $+0.029$ & 0.81 \\
\rowcolor{PanelARow}
S$\rightarrow$L every epoch & $33.465\pm0.097$ & $+0.654$ & $5.5{\times}10^{-5}$ \\
\rowcolor{PanelARow}
L$\rightarrow$S every epoch & $35.723\pm0.136$ & $+2.911$ & $2.9{\times}10^{-8}$ \\
\addlinespace[2.5pt]
\rowcolor{PanelB}
\multicolumn{4}{@{}l}{\textbf{B. Persistent length grouping}} \\
\rowcolor{PanelBRow}
Epoch-wise shuffle & $31.857\pm0.075$ & -- & -- \\
\rowcolor{PanelBRow}
Dynamic grouped & $33.075\pm0.276$ & $+1.219$ & $8.9{\times}10^{-6}$ \\
\rowcolor{PanelBRow}
Static grouped & $33.176\pm0.416$ & $+1.319$ & $7.6{\times}10^{-5}$ \\
\bottomrule
\end{tabular}
\end{table}

Table~\ref{tab:audits} separates batch order from membership. Relative to
Batch shuffle, persistent short-to-long and long-to-short order raise PPL by
0.654 and 2.911. First-epoch short-to-long order changes PPL by only 0.029
($p=0.81$). Neither sorted schedule improves over batch shuffling.
Persistent grouping is also worse whether batches are rebuilt or reused;
their difference is not significant ($p=0.57$). The observed gain is specific
to first-epoch grouping under batch-mean loss.

\Needspace{5\baselineskip}
\subsection{Loss normalization and the grouping effect}
\label{sec:weighting}

First-epoch grouping changes batch token counts and hence the coefficient
$1/N_{\mathcal B}$ of each valid target loss. Token-balanced loss replaces it
with fixed $1/Z$. This intervention changes normalization while keeping the
batch construction policy unchanged.

\begin{table}[!t]
\centering
\caption{Token-balanced intervention on Mimi ($n=8$).}
\label{tab:losscontrol}
\small
\renewcommand{\arraystretch}{1.07}
\setlength{\tabcolsep}{1.35pt}
\begin{tabular}{@{}lccc@{}}
\toprule
\rowcolor{PanelA}
\multicolumn{4}{@{}l}{\textbf{A. Best PPL under each loss}} \\
\rowcolor{TblHead}
Loss & Shuffle & Composition & $\Delta_{\rm comp}$ \\
\rowcolor{PanelARow}
Batch mean & $31.857\pm.075$ & $31.682\pm.083$ & $-0.175$ \\
\rowcolor{PanelARow}
Token balanced & $31.856\pm.077$ & $31.880\pm.066$ & $+0.024$ n.s. \\
\addlinespace[2.5pt]
\rowcolor{PanelB}
\multicolumn{4}{@{}l}{\textbf{B. Effect of switching to token-balanced loss}} \\
\rowcolor{TblHead}
Statistic & Shuffle & Composition & Interaction \\
\rowcolor{PanelBRow}
$\Delta$ PPL & $-0.001$ & $+0.198$ & $+0.199$ \\
\rowcolor{PanelBRow}
$p$ & 0.85 & $3.1{\times}10^{-5}$ & $5.8{\times}10^{-5}$ \\
\bottomrule
\end{tabular}
\end{table}

\begin{figure}[t]
\centering
\includegraphics[width=0.76\columnwidth]{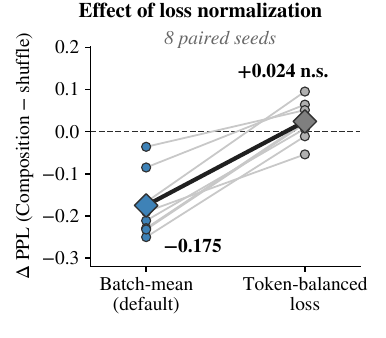}
\caption{Paired composition effect under batch-mean and token-balanced losses
($n=8$).}
\label{fig:tokenbalanced}
\end{figure}

Table~\ref{tab:losscontrol} shows that token balancing affects the two
settings differently. Shuffle changes by only $-0.001$ PPL, whereas
Composition increases by 0.198 PPL. The resulting interaction is 0.199 PPL
($p=5.8\times10^{-5}$). The composition effect changes from $-0.175$ to
$+0.024$ PPL and is no longer statistically significant ($p=0.18$).
Figure~\ref{fig:tokenbalanced} shows an upward shift for all eight paired
seeds. The interaction supports a role for batch-dependent normalization in
Mimi's first-epoch grouping effect.

\Needspace{7\baselineskip}
\subsection{Dependence on sequence-length variation}
\label{sec:tokenizers}

Table~\ref{tab:tokenizers} reports within-tokenizer effects from
Eq.~\ref{eq:factorization}. The coefficient diagnostic uses stored lengths
$c_i$, before shifting. Let $A_{\mathcal B}=\sum_{j\in\mathcal B}c_j$ and
$a_i=1/A_{\mathcal B(i)}$. With $S=\{i:c_i\leq Q_{25}\}$ and
$L=\{i:c_i\geq Q_{75}\}$, computed within the grouped arm,
\par\Needspace{4\baselineskip}
\begin{equation}
\rho_{\rm pre}=\frac{|S|^{-1}\sum_{i\in S}a_i}
{|L|^{-1}\sum_{i\in L}a_i}.
\label{eq:weight_ratio}
\end{equation}
Means are over chunks. Percentile cutoffs are inclusive, so ties can make
the sets overlap. This is a pre-shift diagnostic; training uses
$1/\sum_j(c_j-1)$. Absolute PPL is not compared across tokenizers.

\begin{table}[!t]
\centering
\caption{Cross-tokenizer results ($n=8$). $\rho_{\rm pre}$ is the pre-shift diagnostic in Eq.~\ref{eq:weight_ratio}.}
\label{tab:tokenizers}
\small
\renewcommand{\arraystretch}{1.05}
\setlength{\tabcolsep}{1.45pt}
\begin{tabular}{@{}lccccc@{}}
\toprule
\rowcolor{TblHead}
Tokenizer & CV & Shuffle PPL & $\Delta_{\rm comp}$ & $\Delta_{\rm order}$ & $\rho_{\rm pre}$ \\
\midrule
\rowcolor{KeyRow}
Mimi & 0.309 & $31.857\pm.075$ & $-0.175$ & $+0.120$ & $2.87\times$ \\
SpeechTok. & 0.080 & $4.652\pm.007$ & $-0.001$ & $+0.001$ & -- \\
EnCodec & 0.026 & $30.756\pm.044$ & $+0.015$ & $-0.009$ & $1.00\times$ \\
\bottomrule
\end{tabular}
\end{table}

Only Mimi shows a significant first-epoch composition gain under batch-mean
loss. Its chunk-length CV is 0.309, compared with 0.080 for SpeechTokenizer
and 0.026 for EnCodec. The grouped pre-shift ratio is 2.87 for Mimi and 1.00
for EnCodec. With many full-length chunks, inclusive percentile groups need
not represent separate quarters. These statistics and the loss intervention
support a role for length variation in batch-dependent normalization.
The protocol tests ordering with fixed batches, then tests the remaining
grouping gain by changing the normalization rule.

\section{Conclusion}
\label{sec:conclusion}

In the tested speech-token language model settings, short-to-long ordering
does not improve perplexity when batch composition and token exposure are
controlled. First-epoch length grouping lowers perplexity for Mimi under
batch-mean loss, but this gain is not observed with token-balanced
normalization. The interaction supports a role for batch-dependent loss
normalization. Cross-tokenizer comparisons are consistent with a role of
chunk-length variation. This work provides a systematic analysis protocol
for separating presentation order, token exposure, and loss normalization
in length-based training.

\clearpage
\small
\bibliographystyle{IEEEbib}
\bibliography{strings,refs}

\end{document}